\documentclass{article}

\usepackage[preprint]{neurips_2020}
\usepackage{natbib}

\usepackage[utf8]{inputenc} 
\DeclareUnicodeCharacter{FB01}{fi}
\usepackage[T1]{fontenc}    
\usepackage{url}            
\usepackage{booktabs}       
\usepackage{amsfonts}       
\usepackage{nicefrac}       
\usepackage{microtype}      
\usepackage{color}
\usepackage{mathtools}
\usepackage{bbm}
\usepackage{bm}
\usepackage{braket}
\usepackage{xspace}

\usepackage{hyperref}
\hypersetup{linkcolor=blue, citecolor=black, urlcolor=blue,pagebackref=true,breaklinks=true,letterpaper=true,colorlinks,bookmarks=false}

\usepackage{amsmath,amsthm,commath}
\usepackage[english]{babel}

\title{DeeperGCN: All You Need to Train Deeper GCNs\small\\\url{https://www.deepgcns.org}}

\author{Guohao Li\thanks{equal contribution} \quad Chenxin Xiong \footnotemark[1] \quad Ali Thabet\quad Bernard Ghanem\\
		Visual Computing Center,~ KAUST \\ ~ Thuwal,~ Saudi Arabia \\
		{\tt\footnotesize \{guohao.li, chenxin.xiong, ali.thabet, bernard.ghanem\}@kaust.edu.sa}
		}

\newcommand{\etc}{\emph{etc.}\xspace}
\newcommand{\eg}{\emph{e.g.}\xspace}
\newcommand{\ie}{\emph{i.e.}\xspace}

\newcommand{\figLabel}{Figure\xspace}

\newcommand{\secLabel}{Section\xspace}
\newcommand{\tblLabel}{Table\xspace}
\newcommand{\proLabel}{Property\xspace}
\newcommand{\ppsLabel}{Proposition\xspace}
\newcommand{\defLabel}{Definition\xspace}

\newcommand{\mysection}[1]{\vspace{0pt}\noindent\textbf{#1.}}

\newcommand{\savespace}{\vspace{-5pt}}
\definecolor{orange}{rgb}{1,0.5,0}
\definecolor{maroon}{rgb}{0.51,0,0}

\theoremstyle{plain}
\newtheorem{theorem}{Theorem}

\newtheorem{property}[theorem]{Property}
\newtheorem{proposition}[theorem]{Proposition}

\theoremstyle{definition}
\newtheorem{definition}[theorem]{Definition}

\theoremstyle{remark}

\makeatletter
\DeclareRobustCommand\onedot{\futurelet\@let@token\@onedot}
\def\@onedot{\ifx\@let@token.\else.\null\fi\xspace}

\def\eg{\emph{e.g}\onedot} 
\def\ie{\emph{i.e}\onedot} 
 
\def\etc{\emph{etc}\onedot} \def\vs{\emph{vs}\onedot}

\makeatother

\begin{document}

\maketitle

\begin{abstract}
Graph Convolutional Networks (GCNs) have been drawing significant attention with the power of representation learning on graphs. Unlike Convolutional Neural Networks (CNNs), which are able to take advantage of stacking very deep layers, GCNs suffer from vanishing gradient, over-smoothing and over-fitting issues when going deeper. These challenges limit the representation power of GCNs on large-scale graphs. This paper proposes DeeperGCN that is capable of successfully and reliably training very deep GCNs. We define differentiable generalized aggregation functions to unify different message aggregation operations (\eg mean, max). We also propose a novel normalization layer namely MsgNorm and a pre-activation version of residual connections for GCNs. Extensive experiments on Open Graph Benchmark (OGB) show DeeperGCN significantly boosts performance over the state-of-the-art on the large scale graph learning tasks of node property prediction and graph property prediction.
\end{abstract}
\section{Introduction}\label{sec:intro}
The rise of availability of non-Euclidean data has recently shed interest into the topic of Graph Convolutional Networks (GCNs). GCNs provide powerful deep learning architectures for unstructured data, like point clouds and graphs. GCNs have already proven to be valuable in several applications including predicting individual relations in social networks \citep{social_tang2009relational}, modelling proteins for drug discovery \citep{chem_zitnik2017predicting,chem_wale2008comparison}, enhancing predictions of recommendation engines \citep{rec_monti2017geometric,rec_ying2018graph}, and efficiently segmenting large point clouds \citep{wang2018dynamic,li2019deepgcns}. Recent works have looked at frameworks to train deeper GCN architectures \citep{li2019deepgcns,li2019deepgcns_journal}. These works demonstrate how increased depth leads to state-of-the-art performances on tasks like point cloud classification and segmentation, and protein interaction prediction. The power of these deep models will become more evident with the introduction of large scale graph datasets. Recently, the introduction of the Open Graph Benchmark (OGB) \citep{hu2020open} made available an abundant number of large scale graph datasets. OGB provides graph datasets for tasks like \textit{node classification}, \textit{link prediction}, and \textit{graph classification}.

Graph convolutions in GCNs are based on the notion of message passing \citep{gilmer2017neural}. At each GCN layer, node features are updated by passing information from its connected (neighbor) nodes. To compute a new node feature in message passing, we first combine information from the node and its neighbors through an aggregation function. We then update the node feature by passing the aggregated value through a differentiable function. Given the nature of graphs, aggregation functions must be permutation invariant. This invariance property guarantees the invariance/equivariance to isomorphic graphs \citep{battaglia2018relational,xu2018powerful}. Popular choices for aggregation functions include \emph{mean} \citep{kipf2016semi}, \emph{max} \citep{hamilton2017inductive}, and \emph{sum} \citep{xu2018powerful}. Recent works suggest that different aggregations have different impact depending on the task. For example, \emph{mean} and \emph{sum} seem suitable choices for tasks like node classification \citep{kipf2016semi}, while \emph{max} is favorable for dealing with 3D point clouds \citep{qi2017pointnet,wang2019dynamic}. Currently, the mechanisms to choose a suitable aggregation function are unclear, and all works rely on empirical analysis.

To complement the efficacy of aggregation functions in GCNs, recent work in \cite{li2019deepgcns} borrowed concepts from 2D CNNs to train very deep GCNs. Specifically, the authors adapt residual connections, dense connections, and dilated convolutions to GCN pipelines. These modules  both solve the vanishing gradient problem, as well as increase the receptive field of GCNs. These new modules are integrated with common ReLU and batch normalization modules. Equipped with these new models, GCNs can train with more than 100 layers. With the introduction of large scale graph datasets in OGB, it is unclear how modules like skip connections and commonly used aggregation functions behave in such large graphs. Despite their potential \citep{kipf2016semi,hamilton2017inductive,veli2018gat,xu2018graph}, it is unclear whether these modules are the ideal choices for handling large scale graph datasets.

In this work, we analyze the performance of GCNs on large scale graphs. To enhance the effect of aggregation functions, we propose a novel \textit{Generalized Aggregation Function} suited for graph convolutions. We show how our function covers all commonly used aggregations. Moreover, we can tune the parameters of our general aggregations to come up with customized functions for different tasks. Our generalized aggregation function is fully differentiable and can also be learned  in an end-to-end fashion. Additionally, we also look at the effect of skip connections in various large scale graph datasets. We show how by modifying current GCN skip connections and introducing a novel graph normalization layer, we can enhance the performance in several benchmarks. All in all, our proposed work presents a suite of techniques to train GCNs on diverse large scale datasets. We show how combining our generalized aggregations, modified skip connection, and graph normalization, can achieve state-of-the-art (SOTA) performance in most OGB benchmarks. We summarize our contributions as follows:

\begin{enumerate}
    \item We propose a novel \textit{Generalized Aggregation Function}. This new function is suitable for GCNs, as it enjoys a permutation invariant property. We show how our generalized aggregation covers commonly used functions such as \emph{mean} and \emph{max} in graph convolutions. Additionally, we show how its parameters can be tuned to improve the performance of diverse GCN tasks. Finally, and since our new function is fully differentiable, we show how its parameters can be learned in an end-to-end fashion.
    
    \item We further enhance the power of GCNs by exploring a modified graph skip connections as well as a novel graph normalization layer. We show in our experiments how each of these new additions enhances the performance of GCNs for large scale graphs.
    
    \item We run extensive experiments on four datasets in the Open Graph Benchmark (OGB), where our new tools achieve impressive results in all tasks. In particular, we improve current state-of-the-art performance by $7.8\%$, $0.2\%$, $6.7\%$ and $0.9\%$ on ogbn-proteins, ogbn-arxiv, ogbg-ppa and ogbg-molhiv, respectively.
\end{enumerate}
\section{Related Work}
\label{sec:related}
\mysection{Graph Convolutional Networks (GCNs)} 
Current GCN algorithms can be divided into two categories: spectral-based and spatial-based.
Based on spectral graph theory, \citet{bruna2013spectral} firstly developed graph convolutions using the Fourier basis of a given graph in the spectral domain. Later, many methods are proposed to apply improvements, extensions, and approximations on spectral-based GCNs \citep{kipf2016semi, defferrard2016convolutional, henaff2015deep, levie2018cayleynets, li2018adaptive}. On the other hand, spatial-based GCNs \citep{hamilton2017inductive, monti2017geometric, niepert2016learning, gao2018large, xu2018powerful, veli2018gat} define graph convolution operations directly on the graph, by aggregating the information from neighbor nodes.
To address the scalability issue of GCNs on large-scale graphs, there are mainly two categories of scalable GCN training algorithms: sampling-based \citep{hamilton2017inductive, chen2018fastgcn, li2018adaptive, chen2018stochastic, graphsaint-iclr20} and clustering-based \citep{chiang2019cluster}.

\mysection{Aggregation Functions for GCNs} GCNs update a node's feature vector through aggregating feature information from its neighbors in the graph. Many different neighborhood aggregation functions that possess a permutation invariant property have been proposed \citep{hamilton2017inductive, veli2018gat, xu2018powerful}. Specifically, \citet{hamilton2017inductive} examine mean, max, and LSTM aggregators, and they empirically find that max and LSTM achieve the best performance. Graph attention networks (GATs) \citep{veli2018gat} employ the attention mechanism \citep{DBLP:journals/corr/BahdanauCB14} to obtain different and trainable weights for neighbor nodes by learning the attention between their feature vectors and that of the central node. Thus, the aggregator in GATs operates like a learnable weighted mean. Furthermore, \citet{xu2018powerful} propose a GCN architecture, denoted Graph Isomorphism Network (GIN), with a sum aggregation that is able to have as large discriminative power as the Weisfeiler-Lehman (WL) graph isomorphism test \citep{weisfeiler1968reduction}.

\mysection{Training Deep GCNs} Despite the rapid and fruitful progress of GCNs, most previous art employ shallow GCNs. Several works attempt different ways of training deeper GCNs \citep{hamilton2017inductive,2017arXiv170201105A,highway,DBLP:conf/icml/XuLTSKJ18}. All these works are however limited to 10 layers of depth before GCN performance would degrade.
Inspired by the benefit of training deep CNN-based networks \citep{he2016deep, huang2017densely, YuKoltun2016}, DeepGCNs \citep{li2019deepgcns} propose to train very deep GCNs (56 layers) 
by adapting residual/dense connections (ResGCN/DenseGCN) and dilated convolutions to GCNs. DeepGCN variants achieve state-of-the art results on S3DIS point cloud semantic segmentation \citep{2017arXiv170201105A} and the PPI dataset.
A further obstacle to train deeper GCNs is over-smoothing, first pointed out by \citet{li2018deeper}. Recent works focus on addressing this phenomenon \citep{klicpera_predict_2019,rong2020dropedge,zhao2020pairnorm}. 
\citet{klicpera_predict_2019} proposes a PageRank-based message passing mechanism, involving the root node in the loop. Alternatively, DropEdge \citep{rong2020dropedge} proposes randomly removing edges from the graph, and PairNorm \citep{zhao2020pairnorm} develops a normalization layer. Their works are however still limited to small scale datasets.
\section{Representation Learning on Graphs}
\mysection{Graph Representation}
A graph $\mathcal{G}$ is usually defined as a tuple of two sets $\mathcal{G}=(\mathcal{V}, \mathcal{E})$, where $\mathcal{V} = \set{v_1, v_2, ..., v_N}$ and $\mathcal{E} \subseteq \mathcal{V} \times \mathcal{V}$ are the sets of vertices and edges, respectively. If an edge $e_{ij} = (v_i, v_j) \in \mathcal{E}$, for an undirected graph, $e_{ij}$ is an edge connecting vertices $v_i$ and $v_j$; for a directed graph, $e_{ij}$ is an edge directed from $v_i$ to $v_j$. Usually, a vertex $v$ and an edge $e$ in the graph are associated with vertex features $\mathbf{h}_{v} \in \mathbb{R}^{D}$ and edge features $\mathbf{h}_{e} \in \mathbb{R}^{C}$ respectively.\footnote{In some cases, vertex features or edge features are absent.}

\mysection{GCNs for Learning Graph Representation}
We define a general graph representation learning operator as $\mathcal{F}$, which takes as input a graph $\mathcal{G}$ and outputs a transformed graph $\mathcal{G}^{\prime}$, \ie $\mathcal{G}^{\prime} = \mathcal{F} (\mathcal{G})$. The features or even the topology of the graph can be learned or updated after the transformation $\mathcal{F}$. Typical graph representation learning operators usually learn latent features or representations for graphs such as DeepWalk \citep{perozzi2014deepwalk}, Planetoid \citep{yang2016revisiting}, Node2Vec \citep{grover2016node2vec}, Chebyshev graph CNN \citep{defferrard2016convolutional}, GCN \citep{kipf2016semi}, MPNN \citep{gilmer2017neural}, GraphSage \citep{hamilton2017inductive}, GAT \citep{veli2018gat} and GIN \citep{xu2018powerful}. In this work, we focus on the GCN family and its message passing framework \citep{gilmer2017neural,battaglia2018relational}. To be specific,  message passing based on GCN operator $\mathcal{F}$ operating on vertex $v \in \mathcal{V}$ at the $l$-th layer is defined as follows:

\savespace
\begin{align}
    \label{eq:mvu}
    &\mathbf{m}_{vu}^{(l)} = \bm{\rho}^{(l)}(\mathbf{h}_{v}^{(l)}, \mathbf{h}_{u}^{(l)}, \mathbf{h}_{e_{vu}}^{(l)}), ~u \in \mathcal{N}(v) \\
    \label{eq:mv}
    &\mathbf{m}_{v}^{(l)} = \bm{\zeta}^{(l)} (\set{\mathbf{m}_{vu}^{(l)}|u \in \mathcal{N}(v)}) \\
    \label{eq:hv+}
    &\mathbf{h}_{v}^{(l+1)} = \bm{\phi}^{(l)}(\mathbf{h}_{v}^{(l)}, \mathbf{m}_{v}^{(l)}),
\end{align}

where $\bm{\rho}^{(l)}, \bm{\zeta}^{(l)}$, and $\bm{\phi}^{(l)}$ are all learnable or differentiable functions for \emph{message construction}, \emph{message aggregation}, and \emph{vertex update}  at the $l$-th layer, respectively. For simplicity, we only consider the case where vertex features are updated at each layer. It is straightforward to extend it to edge features. Message construction function $\bm{\rho}^{(l)}$ is applied to vertex features $\mathbf{h}_{v}^{(l)}$ of $v$, its neighbor's features $\mathbf{h}_{u}^{(l)}$, and the corresponding edge features $\mathbf{h}_{e_{vu}}$ to construct an individual message $\mathbf{m}_{vu}^{(l)}$ for each neighbor $u \in \mathcal{N}(v)$. Message aggregation function $\bm{\zeta}^{(l)}$ is commonly a permutation invariant set function that takes as input a countable unordered message set $\set{\mathbf{m}_{vu}^{(l)}|u \in \mathcal{N}(v)}$, where $\mathbf{m}_{vu}^{(l)} \in \mathbb{R}^{D}$; and outputs a reduced or aggregated message $\mathbf{m}_{v}^{(l)} \in \mathbb{R}^{D}$. The permutation invariance of $\bm{\zeta}^{(l)}$ guarantees the invariance/equivariance to isomorphic graphs \citep{battaglia2018relational}. $\bm{\zeta}^{(l)}$ can be a simply symmetric function such as \emph{mean} \citep{kipf2016semi}, \emph{max} \citep{hamilton2017inductive}, or \emph{sum} \citep{xu2018powerful}. Vertex update function $\bm{\phi}^{(l)}$ combines the original vertex features $\mathbf{h}_{v}^{(l)}$ and the aggregated message $\mathbf{m}_{v}^{(l)}$ to obtain transformed vertex features $\mathbf{h}_{v}^{(l+1)}$.

\section{Aggregation functions for GCNs}
\begin{property}[Graph Isomorphic Equivariance]
\label{pro:gie}
If a message aggregation function $\bm{\zeta}$ is permutation invariant to the message set $\set{\mathbf{m}_{vu}|u \in \mathcal{N}(v)}$, then the message passing based GCN operator $\mathcal{F}$ is equivariant to graph isomorphism, \ie for any isomorphic graphs $\mathcal{G}_1$ and $\mathcal{G}_2 = \sigma \star \mathcal{G}_1$, $\mathcal{F}(\mathcal{G}_2) =\sigma \star \mathcal{F}(\mathcal{G}_1)$, where $\star$ denotes a permutation operator on graphs.
\end{property}

The invariance and equivariance properties on sets or GCNs/GNNs have been discussed in many recent works. \citet{zaheer2017deepset} propose DeepSets based on permutation invariance and equivariance to deal with sets as inputs. \citet{maron2019universality} show the universality of invariant GNNs to any continuous invariant function. \citet{keriven2019universal} further extend it to the equivariant case. \citet{maron2018invariant} compose networks by proposed invariant or equivariant linear layers and show their models are as powerful as any MPNN \citep{gilmer2017neural}. In this work, we study permutation invariant functions of GCNs, which enjoy these proven properties.

\subsection{Generalized Message Aggregation Functions} \label{sec:gmaf}
To embrace the nice properties of invariance and equivariance (\proLabel \ref{pro:gie}), many works in the graph learning field tend to use simple permutation invariant functions like \emph{mean} \citep{kipf2016semi}, \emph{max} \citep{hamilton2017inductive} and \emph{sum} \citep{xu2018powerful}. Inspired by the Weisfeiler-Lehman (WL) graph isomorphism test \citep{weisfeiler1968reduction}, \citet{xu2018powerful} propose a theoretical framework and analyze the representational power of GCNs with \emph{mean}, \emph{max} and \emph{sum} aggregators. Although \emph{mean} and \emph{max} aggregators are proven to be less powerful than the WL test in \citep{xu2018powerful}, they are found to be effective on the tasks of node classification \citep{kipf2016semi,hamilton2017inductive} and 3D point cloud processing \citep{qi2017pointnet,wang2019dynamic}. To further go beyond these simple aggregation functions and study their characteristics, we define generalized aggregation functions in the following.

\begin{definition}[Generalized Message Aggregation Functions]
\label{def:gmaf}
We define a generalized message aggregation function $\bm{\zeta}_{x}(\cdot)$ as a 
function that is parameterized by a continuous variable $x$ to produce a family of permutation invariant set functions, \ie $\forall x$, $\bm{\zeta}_{x}(\cdot)$ is permutation invariant to the order of messages in the set $\set{\mathbf{m}_{vu}|u \in \mathcal{N}(v)}$.
\end{definition}

In order to cover the popular \emph{mean} and \emph{max} aggregations into the generalized space, we further define \emph{generalized mean-max aggregation} for message aggregation. It is easy to include \emph{sum} aggregation. For simplicity, we focus on \emph{mean} and \emph{max}.

\begin{definition}[Generalized Mean-Max Aggregation]
\label{def:gmma}
If there exists a pair of $x$, say $x_1$, $x_2$, such that for any message set $\text{lim}_{x\to x_1} \bm{\zeta}_{x}(\cdot) = \text{Mean}(\cdot)$ \footnote{$\text{Mean}(\cdot)$ denotes the arithmetic mean.} and $\text{lim}_{x\to x_2} \bm{\zeta}_{x}(\cdot) = \text{Max}(\cdot)$, then $\bm{\zeta}_{x}(\cdot)$ is a generalized mean-max aggregation function.
\end{definition}

The nice properties of generalized mean-max aggregation functions can be summarized as follows: \textbf{(1)} they provide a large family of permutation invariant aggregation functions; \textbf{(2)} they are continuous and differentiable on $x$ and are potentially learnable; \textbf{(3)} it is possible to interpolate between $x_1$ and $x_2$ to find a better aggregator than \emph{mean} and \emph{max} for a given task. To empirically validate these properties, we propose two families of generalized mean-max aggregation functions based on \defLabel \ref{def:gmma}, namely \emph{SoftMax aggregation} and \emph{PowerMean aggregation}.

\begin{proposition}[SoftMax Aggregation]
\label{pps:softmax}
Given any message set $\set{\mathbf{m}_{vu}|u \in \mathcal{N}(v)}$, $\mathbf{m}_{vu} \in \mathbb{R}^{D}$, $\text{SoftMax\_Agg}_{\beta}(\cdot)$ is a generalized mean-max aggregation function, where $\text{SoftMax\_Agg}_{\beta}(\cdot) = \sum_{u \in \mathcal{N}(v)} \frac{\text{exp}(\beta\mathbf{m}_{vu})}{\sum_{i \in \mathcal{N}(v)}\text{exp}(\beta\mathbf{m}_{vi})}\cdot \mathbf{m}_{vu}$. Here $\beta$ is a continuous variable called an inverse temperature.
\end{proposition}

The SoftMax function with a temperature has been studied in many machine learning areas, \eg Energy-Based Learning \citep{lecun2006tutorial}, Knowledge Distillation \citep{hinton2015distilling} and Reinforcement Learning \citep{gao2017properties}. Here, for low inverse temperatures $\beta$, $\text{SoftMax\_Agg}_{\beta}(\cdot)$ behaves like a mean aggregation. For high inverse temperatures, it becomes close to a max aggregation. Formally, $\text{lim}_{\beta\to 0}\text{SoftMax\_Agg}_{\beta}(\cdot)=\text{Mean}(\cdot)$ and $\text{lim}_{\beta\to \infty}\text{SoftMax\_Agg}_{\beta}(\cdot)=\text{Max}(\cdot)$. It can be regarded as a weighted summation that depends on the inverse temperature $\beta$ and the values of the elements themselves. The full proof of \ppsLabel \ref{pps:softmax} is in the Appendix.

\begin{proposition}[PowerMean Aggregation]
\label{pps:power}
Given any message set $\set{\mathbf{m}_{vu}|u \in \mathcal{N}(v)}$, $\mathbf{m}_{vu} \in \mathbb{R}^{D}_{+}$, $\text{PowerMean\_Agg}_{p}(\cdot)$ is a generalized mean-max aggregation function, where $\text{PowerMean\_Agg}_{p}(\cdot) = (\frac{1}{\left|\mathcal{N}(v)\right|} \sum_{u \in \mathcal{N}(v)} \mathbf{m}_{vu}^{p})^{1/p}$. Here, $p$ is a non-zero, continuous variable denoting the $q$-th power.
\end{proposition}

Quasi-arithmetic mean \citep{kolmogorov1930notion} was proposed to unify the family of mean functions. Power mean is one member of Quasi-arithmetic mean. It is a generalized mean function that includes harmonic mean, geometric mean, arithmetic mean, and quadratic mean. The main difference between \ppsLabel \ref{pps:softmax} and \ref{pps:power} is that \ppsLabel \ref{pps:power} only holds when message features are all positive, \ie $\mathbf{m}_{vu} \in \mathbb{R}^{D}_{+}$. $\text{PowerMean\_Agg}_{p=1}(\cdot)=\text{Mean}(\cdot)$ and $\text{lim}_{p\to \infty}\text{PowerMean\_Agg}_{p}(\cdot)=\text{Max}(\cdot)$. $\text{PowerMean\_Agg}_{p}(\cdot)$ becomes the harmonic or the geometric mean aggregation when $p=-1$ or $p\to0$ respectively. See the Appendix for the proof.

\subsection{GENeralized Aggregation Networks (GEN)} \label{sec:gen}
\mysection{Generalized  Message  Aggregator} Based on the Propositions above, we construct a simple message passing based GCN network that satisfies the conditions in \ppsLabel \ref{pps:softmax} and \ref{pps:power}. The key idea is to keep all the message features to be positive, so that generalized mean-max aggregation functions ($\text{SoftMax\_Agg}_{\beta}(\cdot)$ and $\text{PowerMean\_Agg}_{p}(\cdot)$) can be applied. We define the message construction function $\bm{\rho}^{(l)}$ as follows:

\savespace
\begin{align}
    \label{eq:gen_mc}
    &\mathbf{m}_{vu}^{(l)} = \bm{\rho}^{(l)}(\mathbf{h}_{v}^{(l)}, \mathbf{h}_{u}^{(l)}, \mathbf{h}_{e_{vu}}^{(l)})=\text{ReLU}(\mathbf{h}_{u}^{(l)}+\mathbbm{1}(\mathbf{h}_{e_{vu}}^{(l)})\cdot\mathbf{h}_{e_{vu}}^{(l)})+\epsilon, ~u \in \mathcal{N}(v)
\end{align}

where the $\text{ReLU}(\cdot)$ function is a rectified linear unit \citep{nair2010rectified} that outputs values to be greater or equal to zero, $\mathbbm{1}(\cdot)$ is an indicator function being $1$ when edge features exist otherwise $0$, $\epsilon$ is a small positive constant chosen as $10^{-7}$. As the conditions are satisfied, we can choose the message aggregation function $\bm{\zeta}^{(l)}(\cdot)$ to be either $\text{SoftMax\_Agg}_{\beta}(\cdot)$ or $\text{PowerMean\_Agg}_{p}(\cdot)$.

\mysection{Better Residual Connections}
DeepGCNs \citep{li2019deepgcns} show residual connections \citep{he2016deep} to be quite helpful in training very deep GCN architectures. They simply build the residual GCN block with components following the ordering: $\text{GraphConv}\to\text{Normalization}\to\text{ReLU}\to\text{Addition}$. However, one important aspect that has not been studied adequately is the effect of the ordering of components, which has been shown to be quite important by \citet{he2016identity}. As suggested by \citet{he2016identity}, the output range of the residual function should to be $(-\infty, +\infty)$. Activation functions such as $\text{ReLU}$ before addition may impede the representational power of deep models. Therefore, we propose a pre-activation variant of residual connections for GCNs, which follows the ordering: $\text{Normalization}\to\text{ReLU}\to\text{GraphConv}\to\text{Addition}$. Empirically, we find that the pre-activation version performs better.

\mysection{Message Normalization} In our experiments, we find normalization techniques play a crucial role in training deep GCNs. We apply normalization methods such as BatchNorm \citep{ioffe2015batch} or LayerNorm \citep{ba2016layer} to normalize vertex features. In addition to this, we also propose a \emph{message normalization} (MsgNorm) layer, which can significantly boost the performance of networks with under-performing aggregation functions. The main idea of \emph{MsgNorm} is to normalize the features of the aggregated message $\mathbf{m}_{v}^{(l)} \in \mathbb{R}^{D}$ by combining them with other features during the vertex update phase. Suppose we apply the MsgNorm to a simple vertex update function $\text{MLP}(\mathbf{h}_{v}^{(l)} + \mathbf{m}_{v}^{(l)})$. The vertex update function becomes as follows:

\savespace
\begin{align}
    \label{eq:mn}
    &\mathbf{h}_{v}^{(l+1)} = \bm{\phi}^{(l)}(\mathbf{h}_{v}^{(l)}, \mathbf{m}_{v}^{(l)}) = \text{MLP}(\mathbf{h}_{v}^{(l)} + s \cdot \lVert\mathbf{h}_{v}^{(l)}\rVert_2 \cdot \frac{\mathbf{m}_{v}^{(l)}}{\lVert\mathbf{m}_{v}^{(l)}\rVert_2} )
\end{align}

where $\text{MLP}(\cdot)$ is a multi-layer perceptron and $s$ is a learnable scaling factor. The aggregated message $\mathbf{m}_{v}^{(l)}$ is first normalized by its $\ell_2$ norm and then scaled by the $\ell_2$ norm of $\mathbf{h}_{v}^{(l)}$ by a factor of $s$. In practice, we set the scaling factor $s$ to be a learnable scalar with an initialized value of $1$. Note that when $s = \lVert\mathbf{m}_{v}^{(l)}\rVert_2 / \lVert\mathbf{h}_{v}^{(l)}\rVert_2 $, the vertex update function reduces to the original form.
\section{Experiments}
We propose \emph{GENeralized Aggregation Networks} (GEN), which are comprised of generalized message aggregators, pre-activation residual connections, and message normalization layers. To evaluate the effectiveness of different components, we perform extensive experiments on the \emph{Open Graph Benchmark} (OGB), which includes a diverse set of challenging and large-scale datasets. We first conduct a comprehensive ablation study on the task of node property prediction on the \emph{ogbn-proteins} dataset. Then, we apply our GEN framework on another node property prediction dataset (\emph{ogbn-arxiv}) and two graph property prediction datasets (\emph{ogbg-molhiv} and \emph{ogbg-ppa}).

\subsection{Experimental Setup}
\mysection{PlainGCN}
This baseline model stacks GCN layers with $\{3,7,14,28,56,112\}$ depth and without skip connections. Each GCN layer shares the same message passing operator as GEN except the aggregation function is replaced by $\text{Sum}(\cdot)$, $\text{Mean}(\cdot)$ or $\text{Max}(\cdot)$ aggregation. Layer normalization \citep{ba2016layer} is used in every layer before the activation function $\text{ReLU}$.

\mysection{ResGCN}
Similar to \citet{li2019deepgcns}, we construct ResGCN by adding residual connections to PlainGCN following the ordering: $\text{GraphGonv}\to\text{Normalization}\to\text{ReLU}\to\text{Addition}$.

\mysection{ResGCN+}
We get the pre-activation version of ResGCN by changing the order of residual connections to $\text{Normalization}\to\text{ReLU}\to\text{GraphGonv}\to\text{Addition}$. We denote it as ResGCN+ to differentiate it from ResGCN.

\mysection{ResGEN}
The ResGEN models are designed using the message passing functions described in \secLabel \ref{sec:gen}. The only difference between ResGEN and ResGCN+ is that generalized message aggregators (\ie $\text{SoftMax\_Agg}_{\beta}(\cdot)$ or $\text{PowerMean\_Agg}_{p}(\cdot)$) are used instead of $\text{Sum}(\cdot)$, $\text{Mean}(\cdot)$, or $\text{Max}(\cdot)$. Here we freeze the values of $\beta$ to $ {10}^{n}$, where $n \in \{-3,-2, -1, 0, 1, 2, 3, 4\}$ and $p$ to $\{-1, 10^{-3}, 1, 2, 3, 4, 5\}$.

\mysection{DyResGEN}
In contrast to ResGEN, DyResGEN learns parameters $\beta$ and $p$ \emph{dynamically} for every layer at every gradient descent step. By learning $\beta$ and $p$, we avoid the need to painstakingly searching for the best hyper-parameters. More importantly, DyResGEN can learn aggregation functions to be adaptive to the training process and the dataset. We also study DyResGEN with $\text{MsgNorm}$ layers that learn the norm scaling factor $s$ of aggregated messages.

 \mysection{Datasets}
Traditional graph datasets have been shown to be limited and unable to provide a reliable evaluation and rigorous comparison among methods \citep{hu2020open} due to several reason including their small scale nature, non-negligible duplication or leakage rates, unrealistic data splits, \etc. Consequently, we conduct our experiments on the recently released datasets of \emph{Open Graph Benchmark} (OGB) \citep{hu2020open}, which overcome the main drawbacks of commonly used datasets and thus are much more realistic and challenging. OGB datasets cover a variety of real-world applications and span several important domains ranging from social and information networks to biological networks, molecular graphs, and knowledge graphs. They also span a variety of predictions tasks at the level of nodes, graphs, and links/edges. In this work, experiments are performed on two OGB datasets for node property prediction and two OGB datasets for graph property prediction. We introduce these four datasets briefly. More detailed information about OGB datasets can be found in \citep{hu2020open}.

\mysection{Node Property Prediction} The two chosen datasets deal with protein-protein association networks (ogbn-proteins) and paper citation networks (ogbn-arxiv). \emph{ogbn-proteins} is an undirected, weighted, and typed (according to species) graph containing $132,534$ nodes and $39,561,252$ edges. All edges come with $8$-dimensional features and each node has an 8-dimensional one-hot feature indicating which species the corresponding protein comes from. \emph{ogbn-arxiv} consists of $169,343$ nodes and $1,166,243$ directed edges. Each node is an arxiv paper represented by a $128$-dimensional features and each directed edge indicates the citation direction. For ogbn-proteins, the prediction task is multi-label and ROC-AUC is used as the evaluation metric. For ogbn-arxiv, it is multi-class and evaluated using accuracy.

\mysection{Graph Property Prediction} Here, we consider two datasets, one of which deals with molecular graphs (ogbg-molhiv) and the other is biological subgraphs (ogbg-ppa). \emph{ogbg-molhiv} has $41,127$ subgraphs, while \emph{ogbg-ppa} consists of $158,100$ subgraphs and it is much denser than ogbg-molhiv. The task of ogbg-molhiv is binary classification and that of ogbg-ppa is multi-class classiﬁcation. The former is evaluated by the ROC-AUC metric and the latter is assessed by accuracy.

\mysection{Implementation Details} We first ablate each proposed component on the ogbn-proteins dataset. We then evaluate our model on the other datasets and compare the performances with state-of-the-art (SOTA) methods. Since the ogbn-proteins dataset is very dense and comparably large, full-batch training is infeasible when considering very deep GCNs. We simply apply a random partition to generate batches for both mini-batch training and test. We set the number of partitions to be $10$ for training and $5$ for test, and we set the batch size to $1$ subgraph. A hidden channel size of $64$ is used for all the ablated models. An Adam optimizer with a learning rate of $0.01$ is used to train models for $1000$ epochs. We implement all our models based on PyTorch Geometric \citep{fey2019} and run all our experiments on a single NVIDIA V100 32GB.

\subsection{Results}
\mysection{Effect of Residual Connections} Experiments in \tblLabel \ref{table:resgcn_ablation} show that residual connections significantly improve the performance of deep GCN models. PlainGCN without skip connections does not gain any improvement from increasing depth. Prominent performance gains can be observed in ResGCN and ResGCN+, as models go deeper. Notably, ResGCN+ reaches $0.858$ ROC-AUC with $112$ layers, which is $0.7\%$ higher than the counterpart in ResGCN. The average results show a consistent improvement of ResGCN+ for the three aggregators ($\text{Sum}$, $\text{Mean}$ and $\text{Max}$). This validates the effectiveness of pre-activation residual connections. We also find that the $\text{Max}$ aggregator performs best.

\savespace
\begin{table}[h!]
\caption{Ablation study on residual connections, model depth and aggregators on ogbn-proteins.}
  \label{table:resgcn_ablation}
  \centering
  \begin{tabular}{llllllllll}
  \toprule
         & \multicolumn{3}{c}{PlainGCN}                  & \multicolumn{3}{c}{ResGCN}                    & \multicolumn{3}{c}{ResGCN+}                   \\ 
          \cmidrule(r){2-10}
\#layers & $\text{Sum}$ & $\text{Mean}$ & $\text{Max}$   & $\text{Sum}$ & $\text{Mean}$ & $\text{Max}$   & $\text{Sum}$ & $\text{Mean}$ & $\text{Max}$   \\ \midrule
3        & 0.824        & 0.793         & \textbf{0.834} & 0.824        & 0.786         & 0.824          & 0.830        & 0.792         & 0.829          \\
7        & 0.811        & 0.796         & 0.823          & 0.831        & 0.803         & 0.843          & 0.841        & 0.813         & 0.845          \\
14       & 0.821        & 0.802         & 0.824          & 0.843        & 0.808         & 0.850          & 0.840        & 0.813         & 0.848          \\
28       & 0.819        & 0.794         & 0.825          & 0.837        & 0.807         & 0.847          & 0.845        & 0.819         & 0.855          \\
56       & 0.824        & 0.808         & 0.825          & 0.841        & 0.813         & \textbf{0.851} & 0.843        & 0.810         & 0.853          \\
112      & 0.823        & 0.810         & 0.824          & 0.840        & 0.805         & \textbf{0.851} & 0.853        & 0.820         & \textbf{0.858} \\ \midrule
avg.     & 0.820        & 0.801         & 0.826          & 0.836        & 0.804         & 0.844          & 0.842        & 0.811         & \textbf{0.848} \\
    \bottomrule
  \end{tabular}
\end{table}

\mysection{Effect of Generalized Message Aggregators} In \tblLabel \ref{table:resgcn_softmax} and \tblLabel \ref{table:resgcn_power_mean}, we examine $\text{SoftMax\_Agg}_{\beta}(\cdot)$ and $\text{PowerMean\_Agg}_{p}(\cdot)$ aggregators, respectively. As both of them are \emph{generalized mean-max aggregations},  they can theoretically achieve a performance that is at least as good as $\text{Mean}$ and $\text{Max}$ aggregations through interpolation. For $\text{SoftMax\_Agg}$, when $\beta={10}^{-3}$, it performs similarly to $\text{Mean}$ aggregation (0.814 \vs 0.811). As $\beta$ increases to ${10}^{2}$, it achieves almost the same performance as $\text{Max}$ aggregation. $112$-layer ResGEN with $\text{SoftMax\_Agg}$ reaches the best ROC-AUC at $0.860$ when $\beta=10^{4}$. For $\text{PowerMean\_Agg}$, we find that $\text{PowerMean\_Agg}$ reaches almost the same ROC-AUC as $\text{Mean}$ when $p=1$ (arithmetic mean). We also observe that all other orders of mean except $p=10^{-3}$ (akin to geometric mean) achieve better performance than the arithmetic mean. $\text{PowerMean\_Agg}$ with $p=5$ reaches the best result. However, due to the numerical issues in PyTorch, we are not able to use larger $p$. These results empirically validate the discussion in \secLabel \ref{sec:gmaf} regarding \ppsLabel \ref{pps:softmax} and \ref{pps:power}.

\savespace
\begin{table}[h]
\caption{Ablation study on ResGEN-$\text{SoftMax\_Agg}$ with different $\beta$ values on ogbn-proteins.}
  \label{table:resgcn_softmax}
  \centering
  \begin{tabular}{lllllllllll}
  \toprule
    \cmidrule(r){3-10}
     \#layers & mean & $10^{-3}$ & $10^{-2}$ & $10^{-1}$ & $1$ & $10$ & $10^{2}$ & $10^{3}$ & $10^{4}$ & max\\
    \midrule
    3 & 0.792 & 0.793 & 0.785 & 0.802 & 0.821 & 0.820  & 0.832 & 0.832 & 0.831  & 0.829\\
    7 & 0.813 & 0.809 & 0.802 & 0.806 & 0.835 & 0.840 & 0.843 & 0.842 & 0.846 & 0.845\\
    14 & 0.813 & 0.810 & 0.814 & 0.816 & 0.833 & 0.845 & 0.847 & 0.850 & 0.845 & 0.848\\
    28 & 0.819 & 0.818 & 0.809 & 0.814 & 0.845 & 0.847 & 0.855 & 0.853 & 0.855 & 0.855\\
    56 & 0.810 & 0.823 & 0.823 & 0.828 & 0.849 & 0.851 & 0.851 & 0.856 & 0.855 & 0.853\\
    112 & 0.820 & 0.829 & 0.824 & 0.824 & 0.844 & 0.857 & 0.855 & 0.859 & \textbf{0.860} & 0.858\\
    \midrule
    avg. & 0.811 & 0.814 & 0.809 & 0.815 & 0.838 & 0.843 & 0.847 & \textbf{0.849} & \textbf{0.849} & 0.848 \\
    \bottomrule
  \end{tabular}
\end{table}

\savespace
\begin{table}[h]
\caption{Ablation study on ResGEN-$\text{PowerMean\_Agg}$ with different $p$ values on ogbn-proteins.}
  \label{table:resgcn_power_mean}
  \centering
  \begin{tabular}{llllllll}
  \toprule
     \#layers & $-1$ & $10^{-3}$ & $1$ & $2$ & $3$ & $4$ & $5$\\
    \midrule
    3 & 0.809 & 0.779 & 0.802 & 0.806 & 0.817 & 0.813 & 0.825\\
    7 & 0.827 & 0.792 & 0.797 & 0.821 & 0.835 & 0.839 & 0.844\\
    14 & 0.828 & 0.783 & 0.814 & 0.839 & 0.835  & 0.845 & 0.844\\
    28 & 0.826 & 0.793 & 0.816 & 0.835 & 0.844  & 0.856 & 0.852\\
    56 & 0.826 & 0.782 & 0.818 & 0.834 & 0.846  & 0.854 & 0.842\\
    112 & 0.825 & 0.788 & 0.824 & 0.840 & 0.842  & 0.852 & \textbf{0.857}\\
    \midrule
    avg. & 0.823 & 0.786 & 0.812 & 0.829 & 0.838 & 0.843 &\textbf{0.845}\\
    \bottomrule
  \end{tabular}
\end{table}

\mysection{Learning Dynamic Aggregators} Many previous works have found that different aggregation functions perform very differently on various datasets \citep{hamilton2017inductive,xu2018powerful}. Trying out every possible aggregator or searching hyper-parameters is computationally expensive. Therefore, we propose DyResGEN to explore the potential of learning dynamic aggregators by learning the parameters $\beta$, $p$, and even $s$ within GEN. In \tblLabel \ref{table:dynamic_res_gen_plus}, we find that learning $\beta$ or $\beta\&s$ of $\text{SoftMax\_Agg}$ boosts the average performance from $0.838$ to $0.850$. Specifically, DyResGEN achieves $0.860$ when $\beta$ is learned. We also see a significant improvement in $\text{PowerMean\_Agg}$. The dash in the table denotes  missing values due to numerical issues. By default, the values of $\beta$, $p$, and $s$ are initialized to $1$ at the beginning of training.

\savespace
\begin{table}[h!]
\caption{Ablation study on DyResGEN on ogbn-proteins. $\beta$, $p$ and $s$ denote the learned parameters.}
  \label{table:dynamic_res_gen_plus}
  \centering
  \begin{tabular}{llll|lll}
  \toprule
   & \multicolumn{3}{c}{$\text{SoftMax\_Agg}
   $} &  \multicolumn{3}{c}{$\text{PowerMean\_Agg}$} \\
   \midrule
     \#layers & $\text{Fixed}$ & $\beta$ & $\beta\&s$ & $\text{Fixed}$ & $p$ & $p\&s$\\
    \midrule
    3 &0.821 & 0.832& 0.837 & 0.802 & 0.818 & 0.838\\
    7 &0.835 &0.846 & 0.848 & 0.797 & 0.841 & 0.851\\
    14 &0.833 & 0.849& 0.851 & 0.814 & 0.840 & 0.849\\
    28 &0.845 &0.852 & 0.853 & 0.816 & 0.847 & \textbf{0.854}\\
    56 &0.849 & \textbf{0.860}& 0.854 & 0.818 & 0.846 & -   \\
    112 & 0.844& 0.858& 0.858 & 0.824 & - & - \\
    \midrule
    avg. & 0.838 &\textbf{0.850} & \textbf{0.850} & 0.812 & 0.838 & \textbf{0.848}\\
    \bottomrule
  \end{tabular}
\end{table}

\mysection{Comparison with SOTA}
We apply our GCN models to three other OGB datasets and compare all results with SOTA posted on OGB Learderboard at the time of this submission. The methods include GCN \citep{kipf2016semi}, GraphSAGE \citep{hamilton2017inductive}, GIN \citep{xu2018powerful}, GIN with virtual nodes, GaAN \citep{zhang18gaan}, and GatedGCN \citep{bresson2018gated}. The provided results on each dataset are obtained by averaging the results from 10 independent runs. It is clear that our proposed GCN models outperform SOTA in all four datasets. In two of these datasets (ogbn-proteins and ogbg-ppa), the improvement is substantial. The implementation details and more intermediate experimental results can be found in the Appendix. 

\savespace
\begin{table}[h!]
\caption{Comparisons with SOTA. * denotes that virtual nodes are used.}
  \label{table:comparision_sotas}
  \centering
  \begin{tabular}{llllllll}
  \toprule
     & GCN  & GraphSAGE & GIN & GIN* & GaAN & GatedGCN & \textbf{Ours}\\
        \midrule
     ogbn-proteins &  0.651 & 0.777  & -- & -- & 0.780 & -- & \textbf{0.858} \\
     ogbn-arxiv &  0.717 & 0.715 & -- & -- & -- & -- & \textbf{0.719}\\
     \midrule
     ogbg-ppa &  0.684 & -- & 0.689 & 0.704 & -- & -- & \textbf{0.771}\\
     ogbg-molhiv &  0.761 & -- & 0.756 & 0.771  & -- &  0.777 & \textbf{0.786} \\
    \bottomrule
  \end{tabular}
\end{table}
\savespace

\section{Conclusion}
\label{sec:conclusion}
To enable training deeper GCNs, we first propose a differentiable generalized message aggregation function, which defines a family of permutation invariant functions. We believe the definition of such a generalized aggregation function provides a new view to the design of aggregation functions in GCNs. We further introduce a new variant of residual connections and message normalization layers. Empirically, we show the effectiveness of training our proposed deep GCN models, whereby we set a new SOTA on four datasets of the challenging Open Graph Benchmark.
\newpage
\mysection{\large{Broader Impact}}
The proposed DeeperGCN models significantly outperform all the SOTA on the biological and chemical graph datasets of OGB. We believe that our models are able to benefit other scientific research to make contributions to accelerating the progress of discovering new drugs, improving the knowledge of the functions and 3D structures of proteins. Despite the good performance of our models, making GCNs go such deep does require more GPUs memory resources and consume more time. Training such deep models will potentially increase the energy consumption. In the future, we will focus on optimizing the efficiency of DeeperGCN models.
\begin{ack}
This work was supported by the King Abdullah University of Science and Technology (KAUST) Office of Sponsored Research through the Visual Computing Center (VCC) funding.
\end{ack}


\bibliographystyle{abbrvnat}
\bibliography{neurips_bib}

\newpage
\begin{appendix}
\section{Discussion on Generalized Message Aggregation Functions}
\begin{figure}[h]
    \centering
    \includegraphics[scale=0.8, trim=30mm 40mm 0mm 0mm]{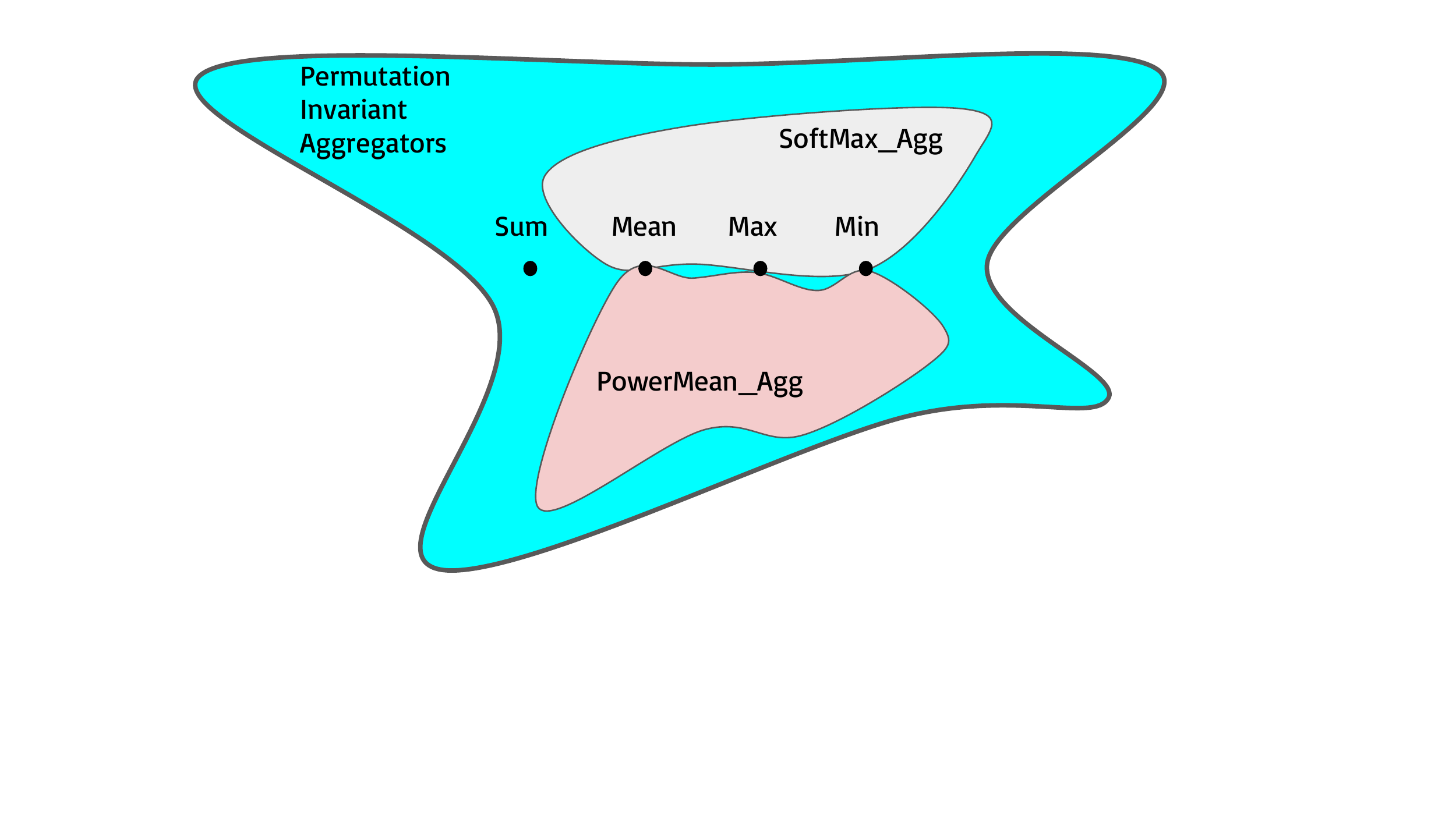}
    \caption{Illustration of Generalized Message Aggregation Functions}
    \label{fig:intro}
\end{figure}

The definition of generalized message aggregation functions help us to find a family of differentiable permutation invariant aggregators. In order to cover the $\text{Mean}$ and $\text{Max}$ aggregations into the function space, we propose two variants of generalized mean-max aggregation functions, \ie $\text{SoftMax\_Agg}_{\beta}(\cdot)$ and $\text{PowerMean\_Agg}_{p}(\cdot)$. They can also be instantiated as a $\text{Min}$ aggregator as $\beta$ or $p$ goes to $-\infty$. We show an illustration of the proposed aggregation functions in \figLabel \ref{fig:intro}. Although we do not generalize the proposed functions to the $\text{Sum}$ aggregation in this work. It could be easily realized by introducing another control variable on the degree of vertices. For instance, we define the form of the function as $\left|\mathcal{N}(v)\right|^{y} \cdot \bm{\zeta}_{x}(\cdot)$ by introducing a variable $y$. We can observe that the function becomes a $\text{Sum}$ aggregation when  $\bm{\zeta}_{x}(\cdot)$ is a $\text{Mean}$ aggregation and $y=1$. We leave it for our future work.

\section{Proof for \ppsLabel \ref{pps:softmax}}
\begin{proof}
Suppose we have $N=\left|\mathcal{N}(v)\right|$. We denote the message set as $\mathbf{M} = \set{\mathbf{m}_1, ..., \mathbf{m}_N}$, $\mathbf{m}_{i} \in \mathbb{R}^{D}$. We first show for any message set, $\text{SoftMax\_Agg}_{\beta}(\mathbf{M}) = \sum_{j=1}^{N} \frac{\text{exp}(\beta\mathbf{m}_{j})}{\sum_{i=1}^{N}\text{exp}(\beta\mathbf{m}_{i})}\cdot \mathbf{m}_{j}$ satisfies \defLabel \ref{def:gmaf}. Let $\rho$ denotes a permutation on the message set $\mathbf{M}$. $\forall \beta \in \mathbb{R}$, for any $\rho \star \mathbf{M} = \set{\mathbf{m}_{\rho(1)}, ..., \mathbf{m}_{\rho(N)}}$, it is obvious that $\sum_{i=\rho(1)}^{\rho(N)}\text{exp}(\beta\mathbf{m}_{i}) = \sum_{i=1}^{N}\text{exp}(\beta\mathbf{m}_{i})$ and $\sum_{j=\rho(1)}^{\rho(N)} \text{exp}(\beta\mathbf{m}_{j})\cdot \mathbf{m}_{j} = \sum_{j=1}^{N} \text{exp}(\beta\mathbf{m}_{j})\cdot \mathbf{m}_{j}$ since the $\text{Sum}$ function is a permutation invariant function. Thus, we have $\text{SoftMax\_Agg}_{\beta}(\mathbf{M})=\text{SoftMax\_Agg}_{\beta}(\rho \star \mathbf{M})$. $\text{SoftMax\_Agg}_{\beta}(\cdot)$ satisfies \defLabel 2.

We now prove $\text{SoftMax\_Agg}_{\beta}(\cdot)$ satisfies \defLabel \ref{def:gmma}, \ie $\text{lim}_{\beta\to 0}\text{SoftMax\_Agg}_{\beta}(\cdot)=\text{Mean}(\cdot)$ and $\text{lim}_{\beta\to \infty}\text{SoftMax\_Agg}_{\beta}(\cdot)=\text{Max}(\cdot)$. For the $k$-th dimension, we have input message features as $\set{m_1^{(k)}, ..., m_N^{(k)}}$. $\text{lim}_{\beta\to 0}\text{SoftMax\_Agg}_{\beta}(\set{m_1^{(k)}, ..., m_N^{(k)}})=\sum_{j=1}^{N} \frac{\text{exp}(\beta m_{j}^{{(k)}})}{\sum_{i=1}^{N}\text{exp}(\beta m_{i}^{{(k)}})}\cdot m_{j}^{{(k)}} = \sum_{j=1}^{N} \frac{1}{N}\cdot m_{j}^{{(k)}} = \frac{1}{N} \sum_{j=1}^{N}\cdot m_{j}^{{(k)}} = \text{Mean}(\set{m_1^{(k)}, ..., m_N^{(k)}})$. Suppose we have $c$ elements that are equal to the maximum value $m^{*}$. When $\beta\to \infty$, we have:
\begin{equation} 
\label{proof:softmax_max}
\frac{\text{exp}(\beta m_{j}^{{(k)}})}{\sum_{i=1}^{N}\text{exp}(\beta m_{i}^{{(k)}})} = \frac{1}{\sum_{i=1}^{N}\text{exp}(\beta (m_{i}^{{(k)}}-m_{j}^{{(k)}}))} = \begin{cases}
1/c &\text{for $m_{j}^{{(k)}} = m^{*}$}\\
0 &\text{for $m_{j}^{{(k)}} < m^{*}$}
\end{cases}
\end{equation}
We obtain $\text{lim}_{\beta\to \infty}\text{SoftMax\_Agg}_{\beta}(\set{m_1^{(k)}, ..., m_N^{(k)}})=c \cdot \frac{1}{c} \cdot m^{*}=m^{*}=\text{Max}(\set{m_1^{(k)}, ..., m_N^{(k)}})$. It is obvious that the conclusions above generalize to all the dimensions. Therefore, $\text{SoftMax\_Agg}_{\beta}(\cdot)$ is a generalized mean-max aggregation function.
\end{proof}

\section{Proof for \ppsLabel \ref{pps:power}}
\begin{proof}
Suppose we have $N=\left|\mathcal{N}(v)\right|$. We denote the message set as $\mathbf{M} = \set{\mathbf{m}_1, ..., \mathbf{m}_N}$, $\mathbf{m}_{i} \in \mathbb{R}^{D}_{+}$. We have  $\text{PowerMean\_Agg}_{p}(\mathbf{M}) = (\frac{1}{N} \sum_{i=1}^{N} \mathbf{m}_{i}^{p})^{1/p}$, $p\neq0$. Clearly, for any permutation $\rho \star \mathbf{M} = \set{\mathbf{m}_{\rho(1)}, ..., \mathbf{m}_{\rho(N)}}$, $ \text{PowerMean\_Agg}_{p}(\rho \star \mathbf{M})= \text{PowerMean\_Agg}_{p}(\mathbf{M})$. Hence, $\text{PowerMean\_Agg}_{p}(\cdot)$ satisfies \defLabel \ref{def:gmaf}. Then we prove $\text{PowerMean\_Agg}_{p}(\cdot)$ satisfies \defLabel \ref{def:gmma} \ie $\text{PowerMean\_Agg}_{p=1}(\cdot)=\text{Mean}(\cdot)$ and $\text{lim}_{p\to \infty}\text{PowerMean\_Agg}_{p}(\cdot)=\text{Max}(\cdot)$. For the $k$-th dimension, we have input message features as $\set{m_1^{(k)}, ..., m_N^{(k)}}$. $\text{PowerMean\_Agg}_{p=1}(\set{m_1^{(k)}, ..., m_N^{(k)}})=\frac{1}{N} \sum_{i=1}^{N}\cdot m_{i}^{{(k)}}=\text{Mean}(\set{m_1^{(k)}, ..., m_N^{(k)}})$. Assume we have $c$ elements that are equal to the maximum value $m^{*}$. When $p\to \infty$, we have:
\begin{align}
\label{proof:powermean_max}
\text{lim}_{p\to \infty}\text{PowerMean\_Agg}_{p}(\set{m_1^{(k)}, ..., m_N^{(k)}}) &= (\frac{1}{N} \sum_{i=1}^{N} (m_{i}^{(k)})^{p})^{1/p} = (\frac{1}{N} (m^{*})^p \sum_{i=1}^{N} (\frac{m_{i}^{(k)}}{m^*})^{p})^{1/p} \\
&= (\frac{c}{N}  (m^{*})^p)^{1/p} \stackrel{m^{*}>0}{=\joinrel=\joinrel=\joinrel=} m^*
\end{align}
We have $\text{lim}_{p\to \infty}\text{PowerMean\_Agg}_{p}(\set{m_1^{(k)}, ..., m_N^{(k)}}) = m^* = \text{Max}(\set{m_1^{(k)}, ..., m_N^{(k)}})$. The conclusions above hold for all the dimensions. Thus, $\text{PowerMean\_Agg}_{p}(\cdot)$ is a generalized mean-max aggregation function.
\end{proof}

\section{Analysis of DyResGEN}
\begin{figure}[h]
    \centering
    \includegraphics[scale=0.3, trim=40mm 10mm 0mm 5mm]{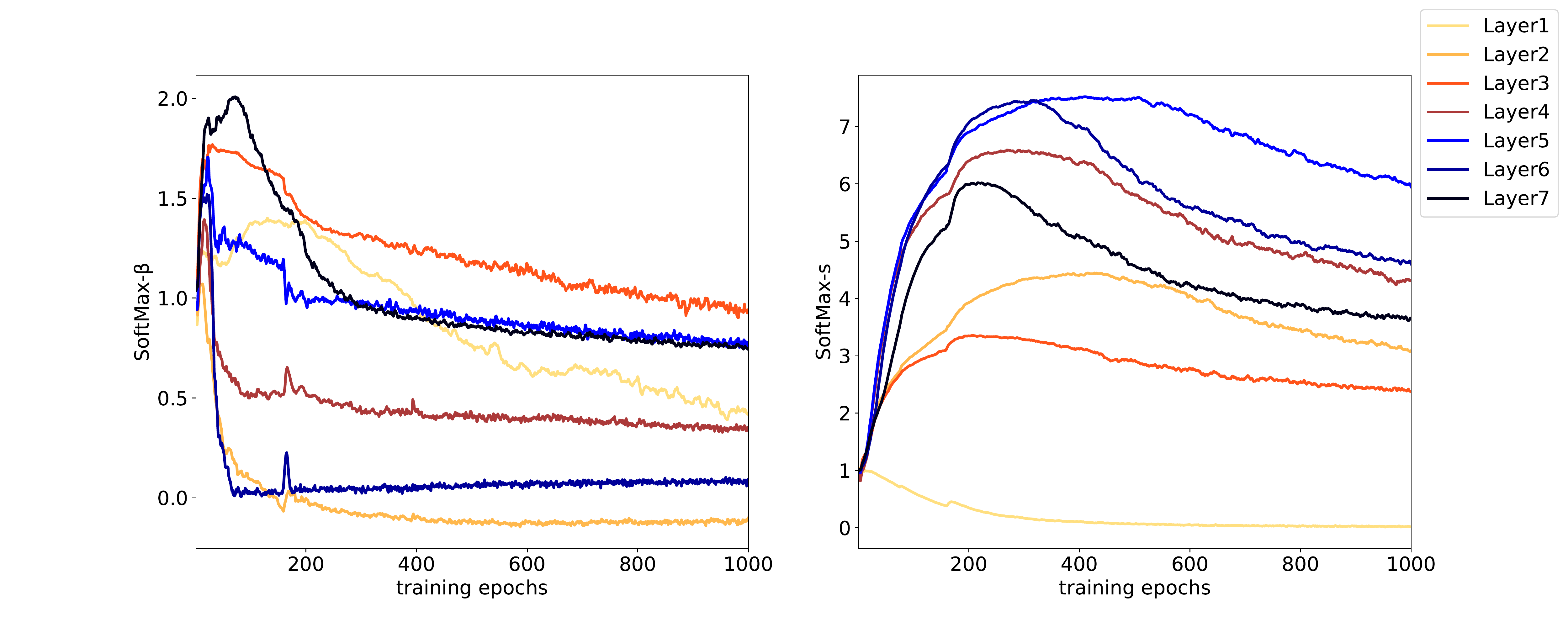}
    \caption{Learning curves of 7-layer DyResGEN with $\text{SoftMax\_Agg}_{\beta}(\cdot)$ and MsgNorm.}
    \label{fig:softmax}
\end{figure}

\begin{figure}[h]
    \centering
    \includegraphics[scale=0.3, trim=40mm 10mm 0mm 5mm]{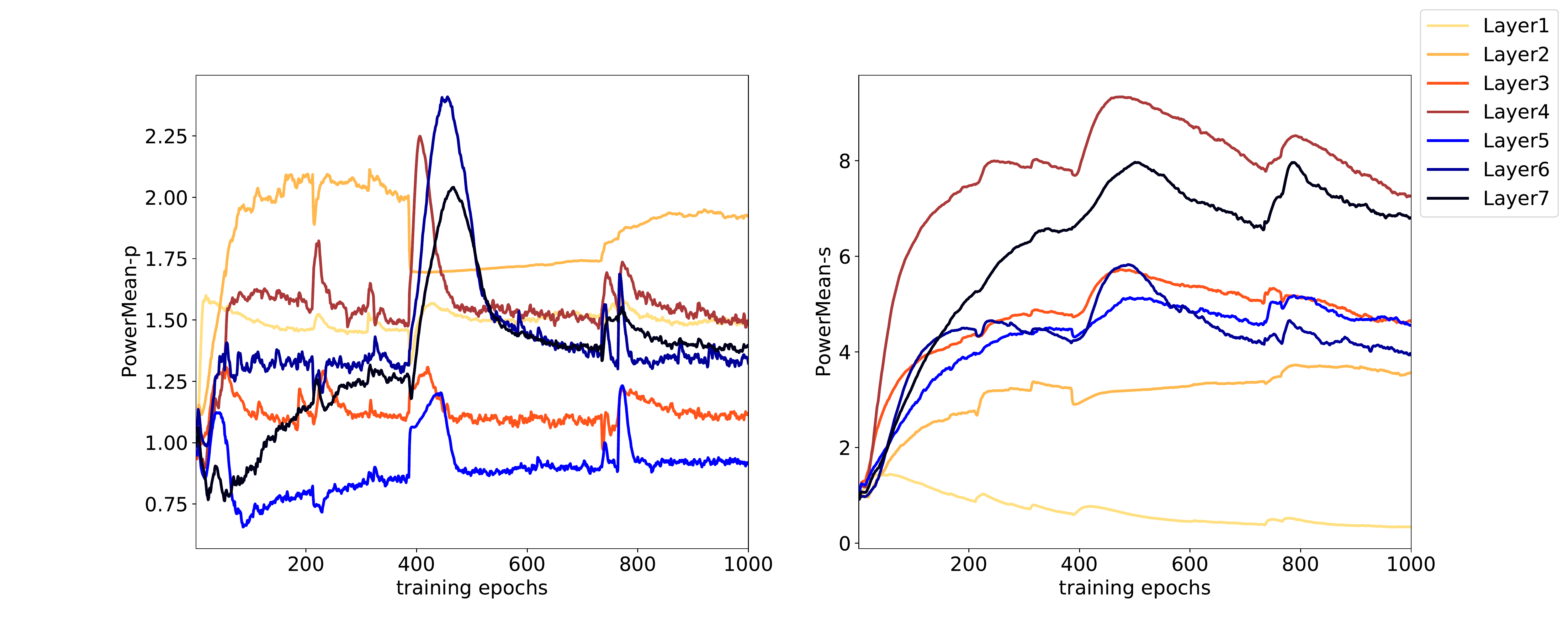}
    \caption{Learning curves of 7-layer DyResGEN with $\text{PowerMean\_Agg}_{p}(\cdot)$ and MsgNorm.}
    \label{fig:power}
\end{figure}

We provide more analysis and some interesting findings of DyResGEN in this section. The experimental results of DyResGEN in this section are obtained on ogbn-proteins dataset. We visualize the learning curves of learnable parameters $\beta$, $p$ and $s$ of $7$-layer DyResGEN with $\text{SoftMax\_Agg}_{\beta}(\cdot)$ aggregator and $\text{PowerMean\_Agg}_{p}(\cdot)$ aggregator respectively. Note that MsgNorm layers with the norm scaling factor $s$ are used. All learnable parameters are initialized as $1$. Dropout with a rate of $0.1$ is used for each layer to prevent over-fitting.
The learning curves of learnable parameters of $\text{SoftMax\_Agg}_{\beta}(\cdot)$ are shown in Figure \ref{fig:softmax}. We observe that both $\beta$ and $s$ change dynamically during the training. The $\beta$ parameters of some layers tend to be stable after $200$ training epochs. Specially, the $6$-th layer learns a $\beta$ to be approximately $0$ which behaves like a $\text{Mean}$ aggregation. For the norm scaling factor $s$, we find that the $s$ of the first layer converges to 0. It indicates that the network learns to aggregation less information from the neighborhood at the first GCN layer, which means a MLP layer could be sufficient for the first layer. Furthermore, we observe that the values of $s$ at deeper layers ($4,5,6,7$) are generally larger than the values at shallow layers ($1,2,3$). It illustrates that the network learns to scale the aggregated messages of the deeper layers with higher scaling factors. Similar phenomena of $\text{PowerMean\_Agg}_{p}(\cdot)$ aggregator are demonstrated in Figure \ref{fig:power}. The change of $p$ shows a larger fluctuation than $\beta$. For the norm scaling factor $s$, the value at the first layer is also learned to be small.

\section{More Details on the Experiments}
In this section, we provide more experimental details on the OGB datasets (ogbn-proteins, ogbn-arxiv, ogbg-ppa and ogbg-molhiv). For a fair comparison with SOTA methods, we provide results on each dataset by averaging the results from 10 independent runs. We provide the details of the model configuration on each dataset. The evaluation results of those 10 independent runs and the corresponding mean and standard deviation are reported in \tblLabel \ref{table:final_results}. The mean and standard deviation are calculated from the raw results. We keep 3 and 4 decimal places for the mean and standard deviation respectively. All models are implemented based on PyTorch Geometric and all experiments are performed on a single NVIDIA V100 32GB.

\begin{table}[h!]
\caption{Results of 10 independent runs on each dataset. The best result among 10 runs is highlighted.}
  \label{table:final_results}
  \centering
  \begin{tabular}{cccc}
  \toprule
    ogbn-proteins & ogbn-arxiv & ogbg-ppa & ogbg-molhiv \\
    \midrule
    0.858 & \textbf{0.721} & 0.768 & 0.774                              \\
0.858                             & 0.719                              & 0.775                     & \textbf{0.805}                     \\
0.857                             & 0.719                              & 0.760                              & 0.789                              \\
0.855                             & 0.718                              & 0.779                              & 0.768                              \\
0.855                             & 0.718                              & 0.773                              & 0.774                              \\
\textbf{0.860}                    & \textbf{0.721}                     & 0.767                              & 0.788                              \\
0.858                             & \textbf{0.721}                     & 0.769                              & 0.794                              \\
0.859                             & 0.718                              & \textbf{0.785}                    & 0.783                              \\
0.859                             & 0.717                              & 0.765                              & 0.786                              \\
\textbf{0.860}                    & 0.720                              & 0.772                              & 0.798                              \\
 \midrule
$\text{0.858} \pm \text{0.0017}$& $\text{0.719} \pm \text{0.0016}$ & $\text{0.771} \pm \text{0.0071}$ & $\text{0.786} \pm \text{0.0117}$ \\

    \bottomrule
  \end{tabular}
\end{table}

\mysection{ogbn-proteins}
For both ogbn-proteins and ogbg-ppa, there is no node feature provided. We initialize the features of nodes through aggregating the features of their connected edges by a $\text{Sum}$ aggregation, \ie $\mathbf{x_i} = {\sum_{j \in \mathcal{N}(i)}}\ \mathbf{e}_{i,j}$, where $\mathbf{x_i}$ denotes the initialized node features and $\mathbf{e}_{i,j}$ denotes the input edge features. We train a $112$-layer DyResGEN with $\text{SoftMax\_Agg}_{\beta}(\cdot)$ aggregator. A hidden channel size of $64$ is used. A layer normalization and a dropout with a rate of $0.1$ are used for each layer. We train the model for $1000$ epochs with an Adam optimizer with a learning rate of $0.01$.

\mysection{ogbn-arxiv}
We train a 28-layer ResGEN model with $\text{SoftMax\_Agg}_{\beta}(\cdot)$ aggregator where $\beta$ is fixed as 0.1. Full batch training and test are applied. A batch normalization is used for each layer. The hidden channel size is $128$. We apply a dropout with a rate of $0.5$ for each layer. An Adam optimizer with a learning rate of $0.01$ is used to train the model for $500$ epochs.

\mysection{ogbg-ppa}
As mentioned, we initialize the node features via a $\text{Sum}$ aggregation. We train a $28$-layer ResGEN model with $\text{SoftMax\_Agg}_{\beta}(\cdot)$ aggregator where $\beta$ is fixed as $0.01$. We apply a layer normalization for each layer. The hidden channel size is set as 128. A dropout with a rate of $0.5$ is used for each layer. We use an Adam optimizer with a learning rate of $0.01$ to train the model for 200 epochs. 

\mysection{ogbg-molhiv} We train a $7$-layer DyResGEN model with $\text{SoftMax\_Agg}_{\beta}(\cdot)$ aggregator where $\beta$ is learnable. A batch normalization is used for each layer. We set the hidden channel size as $256$. A dropout with a rate of 0.5 is used for each layer. An Adam optimizer with a learning rate of 0.01 are used to train the model for 300 epochs.
\end{appendix}

\end{document}